\documentclass[11pt,a4paper]{article}
\usepackage[hyperref]{acl2020}
\usepackage{times}
\usepackage{latexsym}

\usepackage{float}
\usepackage{graphicx}
\usepackage{balance}
\usepackage{siunitx}
\usepackage{booktabs}
\usepackage{colortbl}
\usepackage{amsmath}

\newcommand{\secref}[1]{Section~\ref{sec:#1}}
\newcommand{\figref}[1]{Figure~\ref{fig:#1}}

\usepackage{microtype}

\aclfinalcopy 
\def\aclpaperid{144} 

\title{Towards Multimodal Vision-Language Models \\Generating Non-Generic Text}

\author{Wes Robbins\\
Montana State University\\
\texttt{wesley.robbins10@gmail.com}
\And
Zanyar Zohourianshahzadi \& Jugal Kalita \\
University of Colorado, Colorado Springs \\
\texttt{\{zzohouri,jkalita\}@uccs.edu} } 

\date{}

\begin{document}
\maketitle
\begin{abstract}
   Vision-language models can assess visual context in an image and generate descriptive text. While the generated text may be accurate and syntactically correct, it is often overly general.  To address this, recent work has used optical character recognition to supplement visual information with text extracted from an image. In this work, we contend that vision-language models can benefit from additional information that can be extracted from an image, but are not used by current models. We modify previous multimodal frameworks to accept relevant information from any number of auxiliary classifiers. In particular, we focus on person names as an additional set of tokens and create a novel image-caption dataset to facilitate captioning with person names. The dataset, Politicians and Athletes in Captions (PAC), consists of captioned images of well-known people in context. By fine-tuning pretrained models with this dataset, we demonstrate a model that can naturally integrate facial recognition tokens into generated text by training on limited data. For the PAC dataset, we provide a discussion on collection and baseline benchmark scores.

\end{abstract}

\begin{figure}[h]
    \centering
    \includegraphics[width=.98\linewidth]{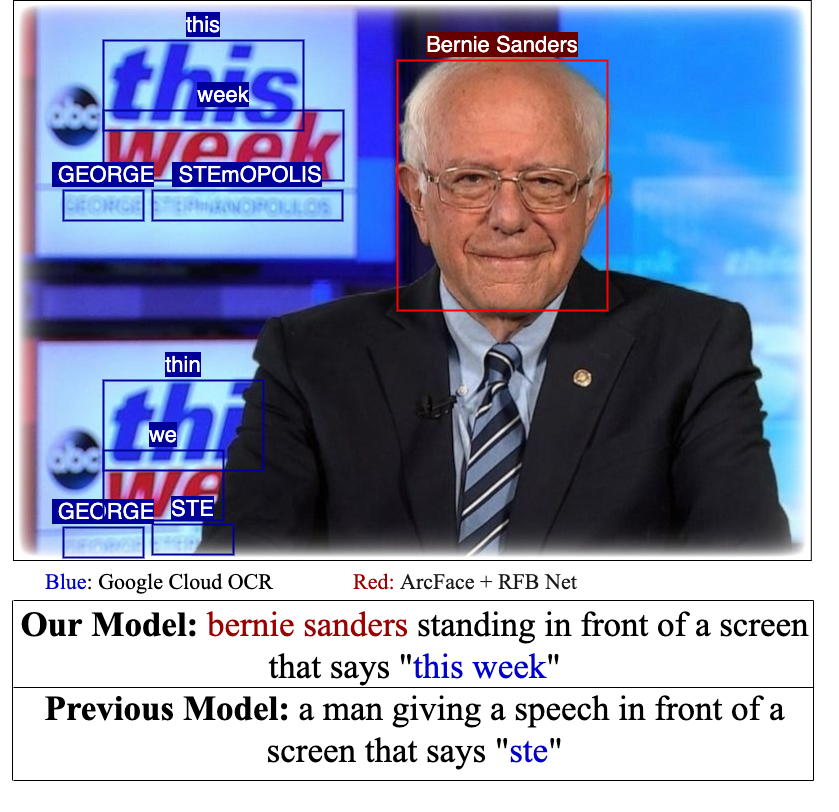}
    \caption{\label{fig:teaser} Our captioning model accepts tokens from several upstream classifiers, learns representations for tokens from different classifiers, and uses each token appropriately. By using the facial recognition token `Bernie Sanders', our model's caption is more informative than previous work which just uses OCR.$^1$}
\end{figure}

\section{Introduction}
\label{sec:intro}
Vision-language models combine deep learning techniques from computer vision and natural language processing to assimilate visual and textual understanding. Such models demonstrate visual and linguistic knowledge by performing tasks such as vision question answering (VQA) and image captioning. There are many applications of these tasks, including aiding the visually impaired by providing scene information and screen reading~\cite{Morris2018RichUsers}.

To perform a vision-language task, a model needs to understand visual context and natural language, and operate in a shared embedding space between the two. 
Approaches in the literature have improved performance by pre-training models for both visual context and  language understanding~\cite{chen2020uniter,Lu_2019_NIPS,su2019vl,li2020oscar,tan2019lxmert}. These models have yielded accurate and semantically appropriate VQAs or captions. However, the text generated from these models are general and overlook content that allow for richer text generation with improved contextualization. For example, they ignore clearly visible text or the presence of well-known individuals.

\footnotetext[1]{\label{firstfootnote}The previous model in \figref{teaser} is M4C Captioner~\cite{Sidorov2020TextCaps:Comprehension} with weights from the M4C repository.}
To improve specificity in generated text, recent work has used optical character recognition (OCR) to incorporate text that appears in images~\cite{zhu2021simple,gao2020multi,mafla2021multi,hu2020iterative,kant2020spatially,wang2021improving,han2020finding,liu2020cascade,yang2021tap}. In many cases, this significantly enhances the usefulness of the generated text~\cite{hu2020iterative}. Such frameworks include OCR as an additional input modality. This results in three modalities for VQA (image, question, and OCR) and two modalities for image captioning (image and OCR). 

While using OCR allows enhancement of some generated text, specific information that exists in human-level description may also come from additional sources. Without proper nouns or other specific vocabulary, the generated text is at the risk of being awkwardly general, demonstrating a lack of shared knowledge that is expected in society. For example in \figref{teaser}, arguably the most relevant content in the image is the presence of a well-known political figure. Consequently, a reasonable description of the image should include the name of the well-known figure, which is `Bernie Sanders' is in this case, instead of generic ``a man". This is notably absent in the caption from the previous model. 

In this work, we propose the \textit{special token approach}, a novel method for integrating tokens from several upstream vision classifiers into image captions.\footnote[2]{While we focus on image captioning, our method could work for integrating non-generic terms into other vision-language tasks such as VQA or visual dialogue which we leave for future work.} We generalize the OCR input modality to accept additional helpful outputs from any number of auxiliary classifiers (\secref{m4c}). We use a rich feature representation for upstream tokens that allows the captioning model to learn to differentiate tokens from different classifiers (\secref{rich}). 

This method potentially allows a model to leverage easily available sophisticated libraries to recognize faces, scene-text, cityscapes, animal species, etc. We refer to all tokens from upstream sources, including OCR tokens, as special tokens. In this work, we focus on using person names and scene-text as example special tokens.

To facilitate using person names in image captions, we create a novel image-caption dataset, Politicians and Athletes in Captions (PAC),  which includes person names in captions in addition to relevant scene-text found on signs, labels, or other entities in the image. PAC has 1,572 images and three captions per image. A discussion on the dataset is provided in \secref{data}. 

By training on PAC in addition to other image-caption datasets, we create a model that can naturally integrate person names into captions. The same model still performs well on previous image captioning benchmarks. 
Evaluation of the methods is available in \secref{exper}.

In summary, this paper makes three primary contributions. The special tokens framework is proposed as a method to incorporate tokens from several external sources into generated text. The PAC image-captioning dataset is collected and baseline results are presented. Lastly, this paper demonstrates the first model in the literature that integrates both facial recognition and OCR into image captioning.

\begin{figure*}[h]
    \centering
    \includegraphics[width=\textwidth]{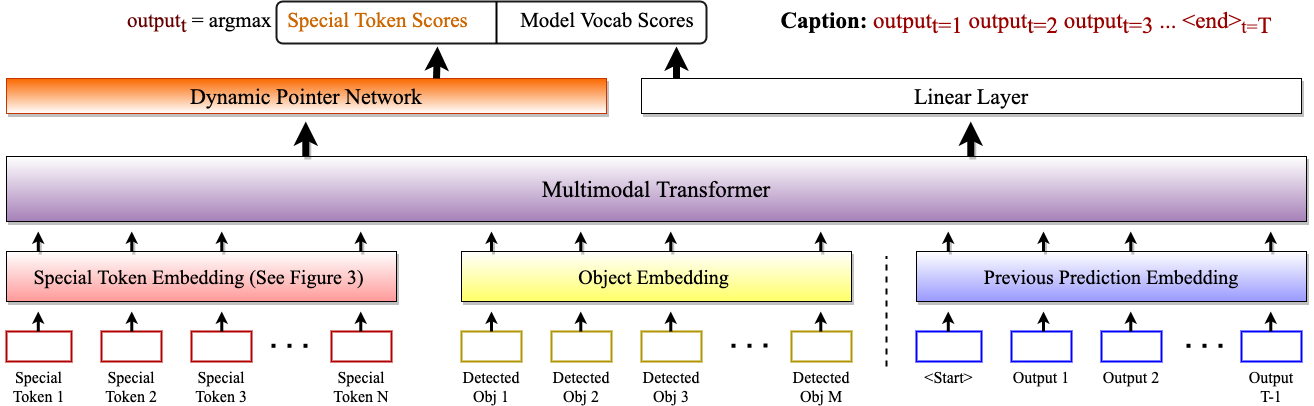}
    \caption{\label{fig:arch} The architecture of the M4C + Special Tokens model. All tokens from upstream classifiers are received by the Special Token modality. The captioning model scores each vocab word and special token at each time step and outputs the highest scoring word. Our method is auto-regressive such that the caption is terminated once an \textit{$<$end$>$} token in generated. Architecture is based on Figure 2 in Hu et al. and updated according to changes outlined in \secref{m4c}.}
\end{figure*}
\section{Related Work}
\label{sec:rw}

The ubiquitous encoder-decoder architecture divides the image captioning task into two parts. The encoder acts as feature extractor and the decoder handles word generation. 
Early deep learning models for image captioning used CNN encoders for feature extraction from the input image as a whole
\cite{kiros2014_1,Karpathy_2014,Vinyals_2015_CVPR}. 
\par
Current models rely on attention \cite{Attention} to generate high-quality image captions.
The seminal image captioning model, Show, Attend and Tell \cite{Xu}, applied attention mechanism on input visual features and the previously generated word (during inference) at each time step for textual caption word generation. 
The majority of current state-of-the-art methods for image captioning and visual question answering benefit from the bottom-up and top-down attention mechanism \cite{Anderson2018Bottom-UpAnswering}.
Bottom-up attention, a hard attention mechanism,
leverages an object detector, Faster R-CNN \cite{FasterRCNN} to detect the most important regions in the image.
Top-down attention, a soft attention mechanism, performs modulation over the set of input visual features from object detection regions.
Following the adoption of bottom-up attention for OCR features \cite{hu2020iterative}, we use the same mechanism to learn to include features obtained from facial recognition. Rather than Faster R-CNN, we use RFBNet \cite{Liu_2018_ECCV} for facial region detection. For facial feature extraction, we use ArcFace \cite{Deng_2019_CVPR} pre-trained on MegaFace dataset \cite{Kemelmacher-Shlizerman_2016_CVPR}.
\par
Several techniques have been proposed to handle OCR tokens in vision-language tasks. The M4C algorithm uses an indiscriminate attention layer followed by a dynamic pointer network~\cite{hu2020iterative}. The SS-Baseline model uses individual attention blocks for each input modality followed by a single fusion encoding layer~\cite{zhu2021simple}. Several approaches have been proposed to better handle spatial information about OCR tokens~\cite{gao2020multi,Gao2020StructuredTextVQA,wang2021improving,kant2020spatially,han2020finding,yang2021tap}. The MMR method utilizes spacial information about objects and scene-text via a graph structure~\cite{mafla2021multi}. TextOCR was introduced as an end-to-end method for identifying OCR tokens~\cite{singh2021textocr}. TAP was introduced as a method to integrate OCR tokens into pre-training.

More similar to our work, Zhao et al. use an upstream classifier as input to a captioning model. They introduce a multi-gated decoder for handling input from external classifiers~\cite{zhao2019informative}. In contrast, we use general OCR and facial recognition classifiers rather than a web entity recognizer as an upstream classifier. Our approach is different from Zhao et al. in that we use bottom-up and top-down attention rather than a standalone CNN for object detection, use a common embedding space rather than a gated decoder for handling multi-modal inputs, and use rich representations (see \secref{rich}) rather than only textual information for handling tokens from upstream classifiers.

MS-COCO \cite{MS_COCO} is a large dataset for common objects in context used for image captioning. Similar to MS-COCO, Flickr30k \cite{flickr30k} is another common dataset used for image captioning.
Google's conceptual captions \cite{sharma2018conceptual} is a vast dataset used for pre-training multitasking vision-language models and fine-tuning them on other vision-language down stream tasks \cite{Lu_2019_NIPS,Lu_2020_CVPR}. The captions in these datasets are generic.
\par
To facilitate use of optical character recognition in the Vision-Language domain, several datasets have been released, including ST-VQA~\cite{Biten_2019_ICCV} for scene text visual question answering and TextCaps~\cite{Sidorov2020TextCaps:Comprehension} for image captioning with reading comprehension. 
Along with the introduction of TextCaps dataset, the M4C model \cite{hu2020iterative} originally used for visual question answering was adopted for image captioning. We modify the M4C model so that it includes bottom-up facial recognition features.
\par

\begin{figure*}[h]
    \centering
    \includegraphics[width=.9\textwidth]{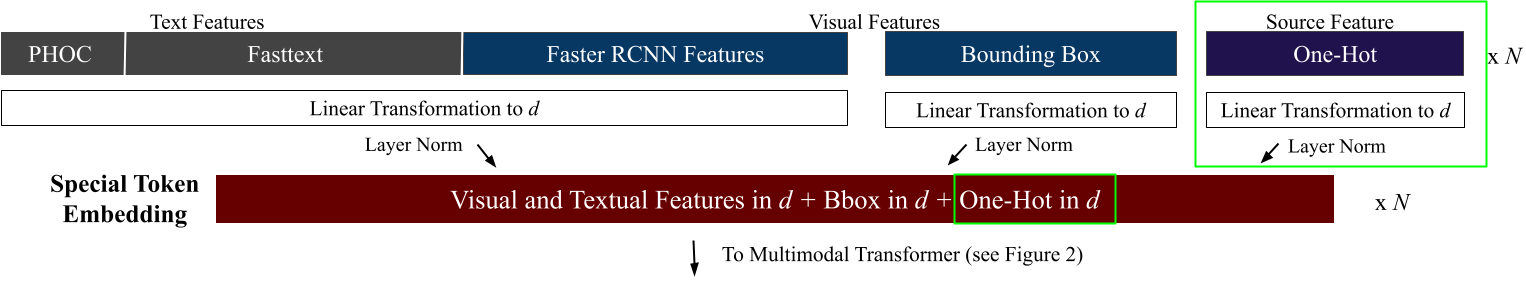}
    \caption{\label{fig:rep} The representation of a special token where $N$ is the number of tokens and $d$ is the dimensionality. We adopt the representation from Hu et al. and add the projected one-hot encoding classifier type feature (highlighted in green box). We are the first to use this representation for facial recognition tokens in addition to OCR tokens. See Equation 2 for more detail.}
\end{figure*}
\newcommand{\lossl}{\mathcal{L}}
\section{Special Tokens}
\label{sec:approach}
We use the term \textit{special token} as a placeholder for extracted relevant information that is identified in an image by upstream sources.  Tokens from upstream classifiers are \textit{special} in that they often are named entities, offering unique descriptors for generic objects. For example in \figref{teaser}, `Bernie Sanders' is not a new object, but rather a special descriptor for an already recognized generic object (i.e. man). Likewise, `this week' is not a generic temporal entity. Instead, it can be used to give more detail about a generic object: a screen that says `this week', referring to a TV show or event called `this week'.

We call our corresponding method for integrating special tokens into image captions the \textit{special token approach}. In our approach, there are two modalities that hold information about an image. The first modality corresponds to generic visual features (yellow box in \figref{arch}) which are responsible for informing the model of general context (all vision-language models have a visual modality). The second modality, special tokens (red box in \figref{arch}), is responsible for informing the model of specific terms that are relevant to the image. The embeddings for the first modality are calculated from visual features from an object detector. The embeddings from the special token modality are calculated from visual feature vectors (Faster-RCNN and a bounding box), textual features (fasttext~\cite{bojanowski2017enriching} and pyramidal histogram of characters (PHOC)~\cite{almazan2014word}), and a source feature (one-hot encoding) as shown in \figref{rep}. Additionally, special tokens are made available for direct copy into generated text which allows for zero-shot inclusion of words not seen prior. This structure has been successful on OCR vision-language datasets. 

The key hypothesis of this paper is that a model can learn to differentiate tokens from separate upstream classifiers. Subsequently, the model can learn to use each token type appropriately in generated text. For example, a caption for the image in \figref{teaser} should neither say ``A screen that says Bernie Sanders" nor should it say `` `this week' standing in front of a screen."  


As mentioned in \secref{intro}, this work demonstrates using two types of special tokens, OCR tokens and facial recognition tokens. We focus our experimentation on learning to integrate facial recognition tokens by training on the PAC dataset. However, any set of words that can be identified by some classification or recognition module can conceivably be a set of special tokens. We leave integration of more upstream vision classifiers for future work.

\subsection{Trade-Offs}
\label{sec:motiv}
The goal of the special token approach is to integrate vocabulary tokens from external sources into generated text. The special tokens approach is based on several following observations. 

1) Different machine learning architectures have been designed to perform well on different tasks. For example, tasks such as OCR detection and facial recognition, benefit from specialized methods that differ from traditional object detection. OCR recognizes and combines characters rather than directly classifying entire words or sentences. In facial recognition, a regression model is trained to output face embeddings which are subsequently compared to embeddings of known individuals. Even in standard classification tasks, significant research is put into fine-tuning architectures to get state-of-the-art results on dataset benchmarks. Such work can be leveraged by a captioning model by using these classifiers as upstream sources.

2) The space of all possible vocabulary tokens, when named entities or proper nouns are included, is intractably large. By appending special tokens to the vocabulary at inference time, the captioning model's vocabulary is prevented from increasing vastly.

3) Using non-generic terms does not always increase the syntactic or semantic complexity of the caption. For example in \figref{teaser}, the name `Bernie Sanders' is a substitution for what can also be a generic term such as `man'. If a captioning model can generate a caption such as `A person standing in front of a screen', the same contextual understanding should be able to generate the caption `Bernie Sanders standing in front of a screen.' The model just needs to know to \textit{use} the named entity `Bernie Sanders'. The special token approach takes advantage of this by allowing the model to learn representations for \textit{types} of special tokens. In \secref{espace} we show that our model learns to represent different token types in  different sections of the embedding space. The model can then implicitly associate sections of the embedding space with related generic objects. 

4) The desired vocabulary may not be constant. For example, after an election cycle, new politicians become commonplace and a captioning model may need to adapt accordingly. The special token approach is highly practical in this sense. The captioning model does not need re-training, only the upstream facial recognition model needs to be updated. 

\subsection{Adopting M4C}
\label{sec:m4c}
We utilize the multimodal multi-copy mesh copy (M4C) model introduced by Hu et al. in order to copy special tokens into generated text~\cite{hu2020iterative}. 
We are the first to utilize this method for tokens other than OCR. Here, we formalize the differences between our captioning model and the M4C captioning model. \figref{arch} provides a corresponding architecture diagram.

The input modalities into the M4C captioning model are object features $\{x_1^{obj},...,x_M^{obj}\}$ for $M$ objects and OCR tokens $\{x_1^{ocr},...,x_N^{ocr}\}$ for $N$ OCR tokens. We generalize OCR tokens to special tokens $st$ such that the inputs are $\{x_1^{obj},...,x_M^{obj}\}$ and $\{x_1^{st},...,x_N^{st}\}$ for $N$ tokens in total. M4C captioner predicts fixed vocab scores $\{y_{1,t}^{voc},...,y_{K,t}^{voc}\}$ where $K$ is a fixed vocabulary size and $t$ is the decoding step, and OCR vocabulary scores $\{y_{1,t}^{ocr},...,y_{N,t}^{ocr}\}$ where $N$ is the number of OCR tokens. The selected word at each time step $w_t = argmax(y_t^{all})$ where $y_t^{all}=\{y_t^{voc}\cup y_t^{ocr}\}$. We substitute $y_t^{st}=\{y_{1,t}^{st},...,y_{N,t}^{st}\}$, where N is the number of special tokens, for $y_t^{ocr}$ such that $y_t^{all}= \{y_t^{voc}\cup y_t^{st}\}$. Special token vocabulary scores $y_{1...N,t}^{st}$ are calculated by combining linear transformations of the decoded output $z_t^{dec}$ and the decoded special token representations $z_n^{st}$ as shown below:
\begin{equation}
    y_{n,t}^{st} = (W^{st}z_n^{st}+b^{st})^T(W^{dec}z_t^{dec}+b^{dec}).
\end{equation}
\subsection{Rich Representations}
\label{sec:rich}
Several types of information may be important for determining if and how a special token should be used in generated text. This may include information about where a special token is located in an image, what the token looks like, or how the token was generated. For example, a known person in the center of an image is more likely to be relevant than a small segment of text found on a sign in the background of an image. Several features are used to richly encode these features of each special token. Hu et al. use visual, spatial, and textual features to calculate OCR tokens embeddings~\cite{hu2020iterative}. We adopt this representation for all special tokens and add an additional source feature to differentiate the upstream classifiers used for identifying special tokens.  A formal description of the special token embedding calculation is described below and a visual representation is provided in \figref{rep}.

Special tokens are represented by a feature vector $x_i^{st}$, where $i = 1...N$.  $x_i^{st}$ incorporates visual features, textual features, and a source feature. The visual features include a bounding box $x_i^{b}$ and a feature vector from an object detector $x_i^{fr}$. Following previous work, we use a pretrained Faster-RCNN with a ResNet backbone to generate $x_i^{fr}$ from the RoI created by the bounding box of the token. The textual features are a fasttext~\cite{bojanowski2017enriching} encoding $x_i^{ft}$ and a pyramidal histogram of characters (PHOC)~\cite{almazan2014word} encoding $x_i^{p}$.  The source feature $x_i^{s}$ is a one-hot encoding between upstream classifiers used for generating special tokens. $x_i^{fr}$, $x_i^{ft}$, and $x_i^{p}$ are concatenated together and projected onto a tuned encoding dimensionality $d$ by a learned linear transformation $W_1$. Additionally, $x_i^{b}$ and $x_i^{s}$ are projected onto $d$ by learned linear transformations $W_2$ and $W_3$. These transformations are trained during the same time as the captioning model. Layer normalization $LN$ is applied to the three $d$ dimensional vectors. $x_i^{spec}$ is a result of element wise addition of these three vectors after layer normalization as shown below:

\begin{equation}
\begin{split}
x_i^{spec}=LN(W_1([x_i^{fr};x_i^{ft};x_i^{p}]))\\ + LN(W_2x_i^b) + LN(W_3x_i^s). 
\end{split}
\end{equation}

\begin{figure*}[h]
\centering
\includegraphics[width=\textwidth]{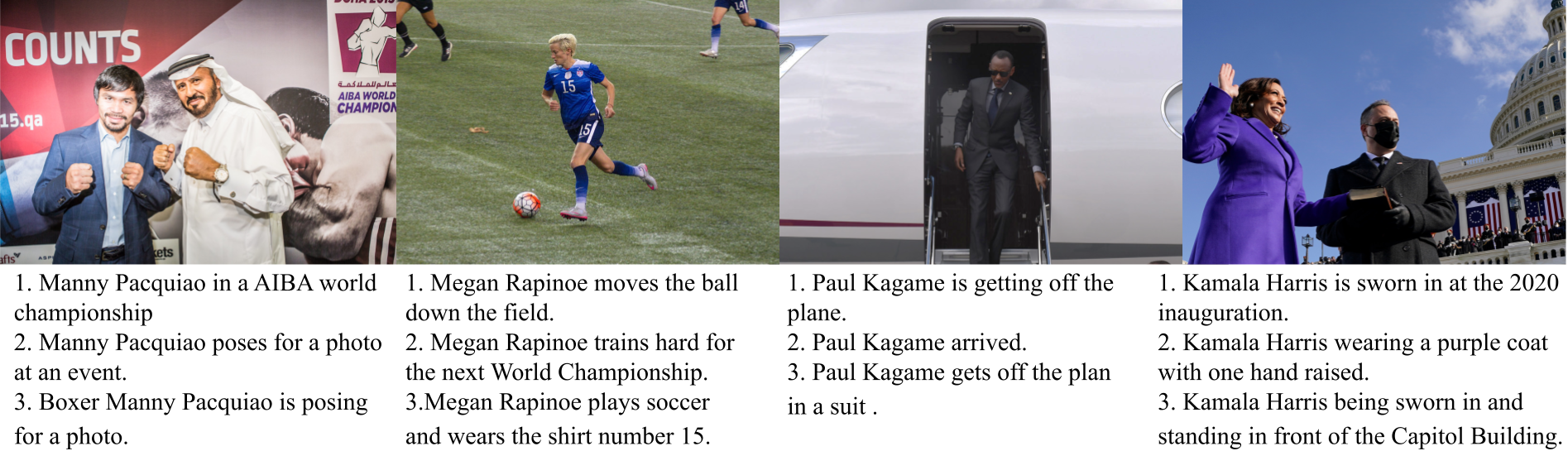}
\caption{\label{fig:samples} Samples from the Politicians and Athletes in Captions dataset}
\end{figure*}

\subsection{Loss}
We do training with decoding binary cross entropy loss $\lossl_{dbce}$ such that the model is supervised at each decoding step $t$ with binary cross entropy $\lossl_{bce}$.
\begin{equation}
\lossl_{dbce} = \sum_{t=1}^{T_{end}} \frac{\lossl_{bce}(t)}{T_{end}}
\end{equation}
where $T_{end}$ is the number of decoding steps before ${<}end{>}$ is predicted from the vocabulary. A maximum number of decoding steps $T_{max}$ is set such that $T_{end} <= T_{max}$.

At each decoding step, sigmoid activation and binary cross entropy are applied uniformly across the fixed model vocabulary of size $K$ and the vector of special tokens of size $N$ such that
\begin{equation}
    \lossl_{bce} = g_n*\log(\sigma(y_n))+(1-g_n)\log(1-\sigma(y_n))
\end{equation}
where $n = 1...{K{+}N}$, $y_n$ is predicted value, and $g_n$ is expected value. 
\section{PAC Dataset}
\label{sec:data}
With this paper we create the Politicians and Athletes in Captions (PAC) dataset. PAC is image-caption dataset consisting of images of well-known individuals in context. PAC includes 1,572 images and three captions per image. Samples from PAC can be seen in \figref{samples} and additional samples can be found in the supplementary materials. 

We create PAC with the goal of studying the use of non-generic vocabulary in image captioning. The non-generic terms emphasized in PAC are person names and OCR tokens. The PAC dataset offers several technical challenges: 1) correctly identifying people in a variety of settings, 2) reasoning about the effect of the \textit{presence} of the individual. If a known person is in a scene, the description of the scene is often based on the known person, and 3) natural integration of a name into a generated caption.


\subsection{Collection}
Images were collected from the Creative Commons image database which are made available under the CC licence. To find individuals for the dataset we searched for `famous athletes' and `famous politicians' and selected 62 individuals. The selected well-known individuals are of various races and sexes and are from many parts of the world. For image collection, we searched for each of the 62 well-known individuals and selected images by manually filtering out duplicates and images without visible faces. 

Annotators were instructed to provide a caption of the image including the name of the individual which was searched for when collecting the image. Other famous individuals who happened to appear in the image may also be mentioned in the captions. Additionally, annotators were instructed to use scene-text if it improved the quality of the caption. These annotation instructions differ from those for caption collection of previous datasets. For example, in the collection of MS-COCO captions, annotators were instructed to \textit{not} use proper nouns~\cite{chen2015microsoft} and annotators for TextCaps were instructed to always use text in the scene~\cite{Sidorov2020TextCaps:Comprehension}. 658 images were captioned by college students and 914 were captioned by Amazon Mechanical Turk. Captions were scanned for grammar and spelling errors.

\begin{figure*}[h]
\centering
\includegraphics[width=\textwidth]{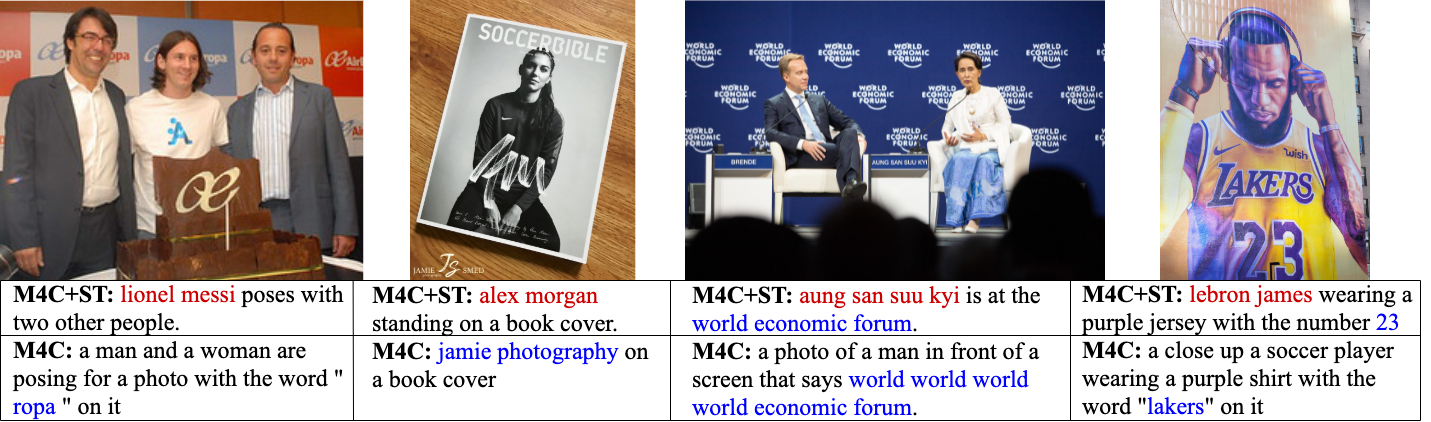}
\caption{\label{fig:qual} Captions generated for PAC test set images. Red words indicate tokens from the face recognition module and blue words indicate tokens from the OCR module. Corresponding metrics found in Table \ref{table:change}.}
\end{figure*}

\subsection{Analysis}
PAC includes images 1,572 images with 3 captions each. All images include at least one famous politician or athlete. Overlap exists in several images. 62 different individuals are in the dataset for an average of 25.2 images per person. 23 of the individuals are politicians while 39 are athletes.

Each caption includes the name of at least one person name in the image. In 66.1\% of images, there is scene text that is recognized by Google Cloud OCR (not all photos have scene text). For 35.9\% of images, at least one of the captions uses scene text (as recognized by Google Cloud OCR). In comparison, 96.9\% of TextCaps images have scene text and 81.3\% of captions use scene text. In the PAC dataset, 96.3\% of the images contain a face region of interest (RoI) that is detected by the RFB Net~\cite{Liu_2018_ECCV}, the face detector we use throughout this work~\cite{Sidorov2020TextCaps:Comprehension}.



\subsection{Limitations}
We identify two primary limitations of the PAC dataset. The dataset with 1,572 images is small relative to similar datasets. 
Due to this PAC cannot represent the breadth of scenes that is found in other datasets. It is recommended to use PAC in conjunction with other dataset in order to mitigate this constraint.

The second primary limitation is narrow scene representation. The dataset is of famous athletes and politicians and therefore overrepresents scenes in which athletes and politicians are photographed. The captions also reflect this bias. For example, the word `suit' is found in 1.82\% of PAC captions while only 0.14\% of TextCaps captions and 0.55\% in MS-COCO. The word `microphone' is found in 1.25\% of PAC captions, 0.11\% in TextCaps, and 0.05\% in MS-COCO. Training on PAC combined with other datasets can mitigate this limitation while still allowing the model to learn to integrate person names as demonstrated in \secref{exper}.
\setlength{\tabcolsep}{9pt}

\begin{table*}[h!]
    
    \centering
    \caption{\label{table:change} Baseline scores on the PAC dataset. Our model (M4C+ST) performs significantly better than a baseline model that does not accept special tokens. For training data, an $\rightarrow$ suggests successive training. A ratio in square brackets represents a sampling ratio for training on both datasets concurrently. We follow previous work and use five common metrics for comparing results.}
    \begin{tabular*}{0.9\textwidth}{@{\extracolsep{\fill}}lllrrrrr}
        \toprule
        \multicolumn{3}{c}{} & 
        \multicolumn{5}{c}{PAC Test Set Metrics}\\
        \cmidrule(lr){4-8}        
        \# & {Model} & {Training} & {B-4} & {M} & {R} & {C} & {S}\\
        \midrule
        {1} & {M4C}   & {TextCaps$\rightarrow$PAC} & 2.1  & 6.4 & 14.3 & 24.6 & 4.3  \\
        {2} & {M4C+ST}& {TextCaps$\rightarrow$PAC} & \textbf{9.1} & \textbf{14.8} & \textbf{30.4} & 102.6 & \textbf{18.7}  \\
        {3} & {M4C+ST}& {PAC,TextCaps[1:8]}        & 8.4 & 14.5 & 30.3 & \textbf{103.7} & 17.5  \\
        {4} & {M4C+ST}& {TextCaps$\rightarrow$PAC,TextCaps[1:1]} & 5.1  & 12.8 & 25.7 & 73.0 & 14.8  \\
        \bottomrule
        \multicolumn{6}{c}{\footnotesize ST: Special Tokens; B-4: BLEU-4; M: METEOR; R: ROUGUE; C: CIDEr, S: SPICE}
    \end{tabular*}
\end{table*}

\section{Experiments}
\label{sec:exper}
In our experiments, we test the special token approach by training on PAC and TextCaps. We present baseline results on PAC. Additionally, we present a visualization for the special token embedding space. 

\subsection{Implementation Details}
For detecting regions in the image with faces, we use RFB Net~\cite{Liu_2018_ECCV}. For facial recognition, we use ArcFace~\cite{Deng_2019_CVPR}. Using ArcFace, we extract facial embeddings for all individuals in the dataset. At inference, we use $l_2$ distance to compare new embeddings to the pre-calculated embeddings. For PAC, ground truth face tokens are known and used during training. The facial recognition model is not used for TextCaps images at training or inference because TextCaps annotators were not instructed to use person names in captions.

We use Google Cloud OCR for extracting OCR tokens. We set a limit at $N = 50$ for the number of special tokens. Face tokens take precedence over OCR tokens if over 50 special tokens are identified. Following previous work, we use a pretrained faster RCNN~\cite{Anderson2018Bottom-UpAnswering} with a ResNet-101 backbone to propose RoIs and extract features for each region. A limit is set at $M=100$ object features. For caption generation, $T_{max}=30$ is the maximum number of decoding steps.

All experiments are performed using either PAC or TextCaps. The captions of these datasets focus on using special tokens (names in PAC, OCR in TextCaps) and are therefore suitable for testing our approach. PAC is broken up into the same 80-20 train-test split for all experiments. We use the specified training and validation sets for TextCaps~\cite{Sidorov2020TextCaps:Comprehension}.

For all training, we use a batch size of 128. We use Adam optimizer with  a learning rate $1e^{-4}$ and learning rate decay of $.1$. 
We use embedding input dimensionality of $d=768$ for inputs to the encoder.

\subsection{Baseline Results}
We first compare our approach (M4C+ST) against the base M4C model. Both models are pretrained on TextCaps, and then trained to convergence on PAC. 
By adding special tokens, we see between 112-334\% percent improvements across metrics on the PAC test set (Table \ref{table:change} Lines 1,2). The vanilla M4C model only has a slight chance of using the correct name which results in poor performance on PAC. 

\figref{qual} shows corresponding qualitative results for the these models. We observe that our model uses person names and OCR tokens appropriately throughout the captions. The right two images demonstrate M4C+ST appropriately switching between model vocabulary, face tokens, and OCR tokens during caption generation. In comparison, the M4C model refers to people generically (i.e `man', `woman',`player') resulting in less informative captions. In the second image, vanilla M4C incorrectly uses `Jamie Photography' (an OCR token found in the bottom left of image) as the name of a person. More qualitative samples from these models can found in the supplementary materials.

In Table \ref{table:change} Lines 3 and 4, we report scores after training on different combinations of PAC and TextCaps. We find that the training procedures from Table \ref{table:change} Lines 2 and 3 are the most effective. Additionally, we find that training on PAC does not degrade performance on TextCaps. Results on the TextCaps dataset can be found in the supplementary materials.

Lastly, we test our model's ability for zero-shot use of tokens from unseen individuals. New images with people not in the PAC dataset are run through our model. Qualitatively, we observe our model is able to integrate unseen individuals into image captions. These samples can be found in the supplementary materials.


\subsection{Special Token Embedding Visualization}
\label{sec:espace}
\begin{figure}[h]
\centering
\includegraphics[width=\linewidth]{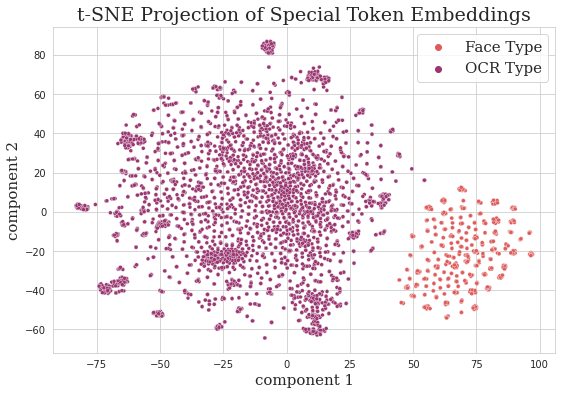}
\caption{\label{fig:tsne} Projection of 768-dimensional special token embeddings into 2d space. Embeddings collected from 314 test images including 703 face tokens and 3,151 OCR tokens.}
\end{figure}

To visualize the embeddings of special tokens, we collect all embeddings during a test set run and plot them with a t-distributed stochastic neighbor embedding (t-SNE). The t-SNE plot shown in \figref{tsne} allows us to visualize the 768-dimensional special token embeddings in 2-dimensions. As previously mentioned, the embeddings are calculated with Equation 2 in \secref{rich}. Both face tokens and OCR tokens go through same learned linear transformation ($W_{1..3}$ from Equation 2),  yet the two different token types are in distinct clusters in the embedding space. This distinction is not known to the model before training, therefore during training the model effectively optimizes $W_{1..3}$ such that embeddings from each token type are meaningfully different for the multimodal transformer. This offers explanation on our observation that our model can use each token type appropriately in generated captions.

\section{Conclusion}
Text generated by vision-language models often lacks specific terms that would be present in human level descriptions or answers. We introduce the special token approach as an adaptable way to introduce non-generic information to a vision-language model. Our method utilizes upstream classifiers to identify information outside of generic context. 
The Politicians and Athletes in Captions dataset consists of image-caption pairs with well-known individuals. By using the special token approach and the PAC dataset, we train a model to integrate person names into image captions. 
Possible improvements to the proposed method include inclusion of more external sources or integration of open-domain knowledge with special tokens. Further progression in this direction could result in captions that are truly interesting, vivid, and useful. 

\section*{Acknowledgement}
The work reported in this paper is supported by the National Science Foundation under Grant No. 2050919. Any opinions, findings and conclusions or recommendations expressed in this work are those of the authors and do not necessarily reflect the views of the National Science Foundation.

\balance
\bibliography{refs}
\bibliographystyle{acl_natbib}
\end{document}